\documentclass[10pt,twocolumn,letterpaper]{article}

\usepackage[pagenumbers]{cvpr} 

\usepackage{graphicx}
\usepackage{amsmath}
\usepackage{amssymb}

\usepackage{booktabs}       
\usepackage{nicefrac}       
\usepackage{microtype}      
\usepackage{amsfonts}       
\usepackage{pifont}
\usepackage{dsfont}
\usepackage{cancel}

\usepackage[dvipsnames]{xcolor} 
\usepackage{xspace}
\usepackage{graphicx}
\usepackage{multirow}
\usepackage{subcaption}
\usepackage{color}
\usepackage{colortbl}
\usepackage{siunitx}
\usepackage{textcomp}

\usepackage{pifont}
\newcommand{\cmark}{\ding{51}}%
\newcommand{\xmark}{\ding{55}}%
\usepackage[misc]{ifsym}

\usepackage{bbm}
\usepackage{listings}
\graphicspath{{figures/}}
\DeclareGraphicsExtensions{.pdf}
\usepackage[accsupp]{axessibility}

\usepackage{amsmath,amsfonts,bm}

\def\eqref#1{equation~\ref{#1}}

\def\1{\bm{1}}

\def\va{{\bm{a}}}
\def\vb{{\bm{b}}}
\def\vc{{\bm{c}}}

\def\vh{{\bm{h}}}

\def\vk{{\bm{k}}}

\def\vq{{\bm{q}}}

\def\vv{{\bm{v}}}

\def\vx{{\bm{x}}}
\def\vy{{\bm{y}}}
\def\vz{{\bm{z}}}

\DeclareMathAlphabet{\mathsfit}{\encodingdefault}{\sfdefault}{m}{sl}
\SetMathAlphabet{\mathsfit}{bold}{\encodingdefault}{\sfdefault}{bx}{n}

\usepackage[pagebackref,breaklinks,colorlinks]{hyperref}

\def\eg{\textit{e.g.}}
\def\ie{\textit{i.e.}}

\def\etal{\textit{et al.}}

\newcommand{\methodname}{\texttt{CIM}\xspace}
\newcommand{\context}{\textit{context}\xspace}
\newcommand{\exemplar}{\textit{exemplar}\xspace}
\newcommand{\exemplars}{\textit{exemplars}\xspace}

\newcommand{\trans}[1]{{#1}^{\ensuremath{\mathsf{T}}}} 

\usepackage[capitalize]{cleveref}
\crefname{section}{Sec.}{Secs.}
\Crefname{section}{Section}{Sections}
\Crefname{table}{Table}{Tables}
\crefname{table}{Tab.}{Tabs.}

\definecolor{aqua}{rgb}{0.0, 1.0, 1.0}
\definecolor{guppiegreen}{rgb}{0.0, 1.0, 0.5}

\begin{document}

\title{
    Correlational Image Modeling for Self-Supervised Visual Pre-Training
}

\author{
Wei Li, Jiahao Xie, Chen Change Loy\\
S-Lab, Nanyang Technological University \\
\tt\small \{wei.l, jiahao003, ccloy\}@ntu.edu.sg\\
}

\twocolumn[{%
\maketitle

\begin{center}
    \centering 
    \includegraphics[width=0.95\linewidth]{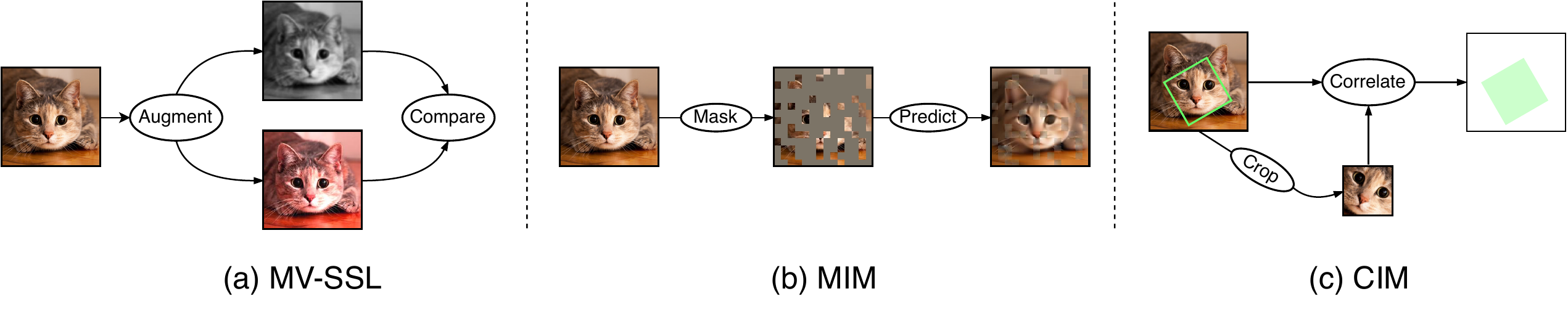}
    \captionof{figure}{
        \textbf{Schematic of pretext tasks} in self-supervised visual pre-training. (\textbf{a}) Multi-View Self-Supervised Learning (MV-SSL) follows an \textbf{\textit{augment-and-compare}} paradigm. (\textbf{b}) Masked Image Modeling (MIM) conducts a \textbf{\textit{mask-and-predict}} pretext task within a single view. (\textbf{c}) Correlational Image Modeling (CIM) formulates a novel \textbf{\textit{crop-and-correlate}} scheme.  
    }
\label{fig:teaser}
\end{center}
}]

\begin{abstract}
We introduce \textbf{C}orrelational \textbf{I}mage \textbf{M}odeling (\methodname), 
a novel and surprisingly effective approach to self-supervised visual pre-training. Our \methodname performs a simple pretext task: we randomly crop image regions (exemplars) from an input image (context) and predict correlation maps between the exemplars and the context. Three key designs enable correlational image modeling as a nontrivial and meaningful self-supervisory task. First, to generate useful exemplar-context pairs, we consider cropping image regions with various scales, shapes, rotations, and transformations. Second, we employ a bootstrap learning framework that involves online and target encoders. During pre-training, the former takes exemplars as inputs while the latter converts the context. Third, we model the output correlation maps via a simple cross-attention block, within which the context serves as queries and the exemplars offer values and keys. We show that \methodname performs on par or better than the current state of the art on self-supervised and transfer benchmarks. Code is available at \url{https://github.com/weivision/Correlational-Image-Modeling.git}.
\end{abstract}
\section{Introduction}
\label{sec:intro}
Recent advances in self-supervised visual pre-training have shown great capability in harvesting meaningful representations from hundreds of millions of---often easily accessible---\textit{unlabeled} images. Among existing pre-training paradigms, Multi-View Self-Supervised Learning (MV-SSL)~\cite{he2020momentum,chen2020simple,chen2020improved,grill2020bootstrap,chen2021exploring,chen2021empirical,caron2021emerging} and Masked Image Modeling (MIM)~\cite{bao2021beit,he2021masked,wei2021masked,xie2021simmim} are two leading methods in the self-supervised learning racetrack,
thanks to their nontrivial and meaningful \textit{self-supervisory pretext tasks}.

MV-SSL follows an \textbf{\textit{augment-and-compare}} paradigm (Figure~\ref{fig:teaser}(a)) -- randomly transforming an input image into two augmented views and then comparing two different views in the representation space. Such an instance-wise discriminative task is rooted in \textit{view-invariant} learning~\cite{tian2020makes}, \ie, changing views of data does not affect the conveyed information.
On the contrary, following the success of Masked Language Modeling (MLM)~\cite{devlin2019bert}, MIM conducts a \textbf{\textit{mask-and-predict}} pretext task within a single view (Figure~\ref{fig:teaser}(b)) -- removing a proportion of random image patches and then learning to predict the missing information. This simple patch-wise
 generative recipe enables Transformer-based deep architectures~\cite{dosovitskiy2020image} to learn generalizable representations from unlabeled images.

Beyond \textbf{\textit{augment-and-compare}} or \textbf{\textit{mask-and-predict}} pretext tasks in MV-SSL and MIM, in this paper, we endeavor to investigate another simple yet effective paradigm for self-supervised visual representation learning. We take inspiration from visual tracking~\cite{yilmaz2006object} in computer vision that defines the task of estimating the motion or trajectory of a target object (\textit{exemplar}) in a sequence of scene images (\textit{contexts}). To cope with challenging factors such as scale variations, deformations, and occlusions, one typical tracking pipeline is formulated as maximizing the correlation between the specific \textit{exemplar} and holistic \textit{contexts}~\cite{bolme2010visual,bertinetto2016fully,valmadre2017end,wang2021multiple}.
Such simple correlational modeling can learn meaningful representations in the capability of both localization and discrimination, thus making it appealing to serve as a promising pretext task for self-supervised learning.

Training a standard correlational tracking model, however, requires access to numerous labeled data, which is unavailable in unsupervised learning. Also, the task goal of visual tracking is intrinsically learning toward one-shot object detection---demanding rich prior knowledge of objectness---while less generic for representation learning. Therefore, it is nontrivial to retrofit supervised correlational modeling for visual tracking into a useful self-supervised pretext task.

Driven by this revelation, we present a novel \textbf{\textit{crop-and-correlate}} paradigm for self-supervised visual representation learning, dubbed as \textbf{C}orrelational \textbf{I}mage \textbf{M}odeling (\methodname).  To enable correlational modeling for effectively self-supervised visual pre-training, we introduce three key designs.
First, as shown in Figure~\ref{fig:teaser}(c), we randomly crop image regions (treated as \textit{exemplars}) with various scales, shapes, rotations, and transformations from an input image (\textit{context}). The corresponding correlation maps can be derived from the exact crop regions directly. 
This simple cropping recipe allows us to easily construct the \textit{exemplar-context} pairs together with ground-truth correlation maps without human labeling cost. Second, we employ a bootstrap learning framework that is comprised of two networks: an online encoder and a target encoder, which, respectively, encode \exemplars and \context into  latent space. This bootstrapping effect works in a way that the model learns to predict the spatial correlation between the updated representation of \exemplars and the slow-moving averaged representation of \context.
Third, to realize correlational learning, we introduce a correlation decoder built with a cross-attention layer and a linear predictor, which computes queries from \context, with keys and values from \exemplars, to predict the corresponding correlation maps.

Our contributions are summarized as follows: \textbf{1)} We present a simple yet effective pretext task for self-supervised
visual pre-training, characterized by a novel unsupervised correlational image modeling framework (\methodname). 
\textbf{2)} We demonstrate the advantages of our \methodname in learning transferable representations for both ViT and ResNet models that can perform on par or better than the current state-of-the-art MIM and MV-SSL learners while improving model robustness and training efficiency.
We hope our work can motivate future research in exploring new useful pretext tasks for self-supervised
visual pre-training.

\section{Related Work}
\label{sec:related_work}
\noindent\textbf{Unsupervised pretext tasks} play the fundamental role in self-supervised representation learning. Beyond \textit{augment-and-compare} and \textit{mask-and-predict}, a series of different unsupervised pretext tasks have been studied in the literature. For instance, Noroozi~\etal~\cite{noroozi2016unsupervised} train a context-free network without human annotation by solving Jigsaw puzzles, further developed in a very recent work~\cite{zhai2022position} by predicting positions from content images.
Bojanowski~\etal~\cite{bojanowski2017unsupervised} propose to learn discriminative features via predicting noise. Gidaris~\etal~\cite{gidaris2018unsupervised} treat the 2D rotation of an image as a supervisory signal. Zhang~\etal~\cite{zhang2019aet} follow this work to predict general affine transformations.
All these initiatives are proven less effective than the state-of-the-art MIM and MV-SSL approaches in large-scale visual pre-training.

\noindent\textbf{Multi-view self-supervised learning} approaches~\cite{wu2018unsupervised,he2020momentum,misra2020self,chen2020simple,chen2020improved,grill2020bootstrap,chen2021exploring,chen2021empirical,caron2021emerging,xie2022delving} are highly successful in learning representations over the past few years. These methods depend on an \textit{augment-and-compare} pretext task that models similarity and dissimilarity between two or more augmented views 
in an embedding space. Thus, MV-SSL greatly relies on data augmentations and Siamese networks~\cite{bromley1993signature}. There have been several general strategies for comparing augmented views. Most contrastive approaches, such as SimCLR~\cite{chen2020simple}, MoCo~\cite{he2020momentum,chen2020improved,chen2021empirical} measure both positive and negative pairs via cosine distance. On the contrary, BYOL~\cite{grill2020bootstrap} and SimSiam~\cite{chen2021exploring} rely only on positive pairs. Beyond contrastive learning, SwAV~\cite{caron2020unsupervised} resorts to online clustering and predicts cluster assignments of different views. In addition, there is another line of research in MV-SSL that extends the main focus of \textit{global} representations to \textit{dense} representations~\cite{pinheiro2020unsupervised,wang2021dense,xie2021propagate,selvaraju2021casting,henaff2021efficient,roh2021spatially,yang2021instance,xiao2021region,xie2021detco,xie2021unsupervised,wei2021aligning,li2022univip,bai2022point}.

\noindent\textbf{Masked image modeling} follows a \textit{mask-and-predict} pretext task, which is inspired by the successful masked language modeling (MLM) approaches in the NLP community, such as BERT~\cite{devlin2019bert} and RoBERTa~\cite{liu2019roberta}. Two key steps can be identified in a typical MIM pipeline: i) \textit{how to mask}, ii) \textit{what to predict}.
In terms of \textit{how to mask}, most MIM approaches, such as BEiT~\cite{bao2021beit}, MAE~\cite{he2021masked} and SimMIM~\cite{xie2021simmim}, extend the \textit{mask-word} recipe in MLM to randomly mask image patches in the spatial domain. Recent works consider other corruptions to replace the normal patch-masking process. For example, Xie~\etal~\cite{xie2022masked} investigate corruption operations (downsample, blur, and noise) in low-level image processing tasks and present a unified \textit{mask-frequency} recipe. Similarly, other degradation forms are studied in Tian~\etal~\cite{tian2022beyond}, including zoom, distortion, and decolorization. Besides, Fang~\etal~\cite{fang2022corrupted} employ an auxiliary generator to corrupt 
the input images. As to \textit{what to predict}, beyond default raw pixels~\cite{he2021masked,xie2021simmim}, several other reconstruction targets are proposed, \eg, hand-crafted or deep features~\cite{wei2021masked}, low or high frequencies~\cite{xie2022masked,liu2022devil}, and discrete tokens~\cite{bao2021beit}. 

\noindent\textbf{Correlational modeling} is the crucial process in visual tracking~\cite{yilmaz2006object}, aiming to predict a dense set of matching confidence for a target object. The seminal work of Correlation Filter~\cite{bolme2010visual} and its end-to-end Siamese-based variants~\cite{bertinetto2016fully,valmadre2017end,li2018high,wang2019fast,wang2021multiple} learn to distinguish targets from background images via convolution (\ie, cross-correlation). Recently, Transformer-based trackers~\cite{chen2021transformer,xie2022correlation,cui2022mixformer,ma2022unified,song2022transformer} employ a cross-attention mechanism to model the correlation between target objects and backgrounds. These promising correlation-based trackers motivate us to investigate the effectiveness of correlational modeling in the context of self-supervised visual pre-training. Notably, some unsupervised and self-supervised trackers~\cite{wang2019unsupervised,wu2021progressive,shen2022unsupervised,lai2020mast,zheng2021learning} normally conduct training on synthetic datasets without labeling. While similarly considering unsupervised or self-supervised training, our work significantly differs from these unsupervised and self-supervised trackers. We will discuss the differences in Section~\ref{subsec:comparison}.

\section{Approach}
\label{sec:methods}
\begin{figure}[!t]
    \centering
    \includegraphics[page=1,width=1.0\linewidth]{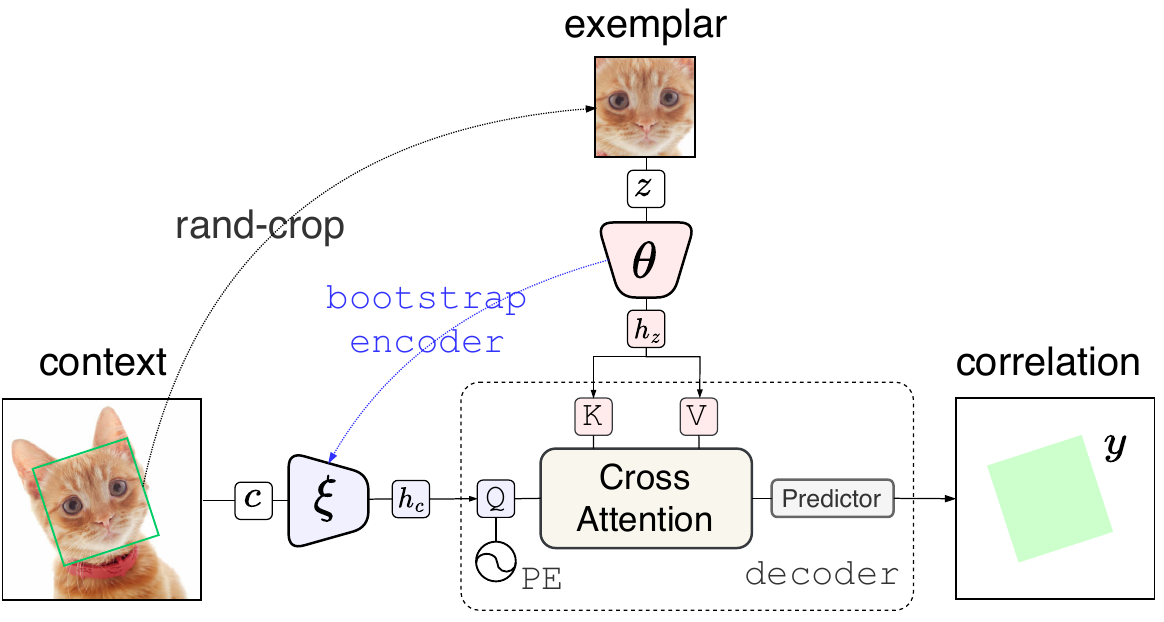}
    \vskip -0.3cm
    \caption{
        The overview of our proposed \methodname pre-training framework. Given an image $\vc$ (\context), we crop a random region $\vz$ (\exemplar) within \context $\vc$. The \context and \exemplar images are separately passed through a target encoder $f_\xi$ and an online encoder $f_\theta$ to obtain latent representations $\vh_c$ and $\vh_z$, which are further fed into a lightweight decoder with a cross-attention layer and a linear predictor to predict the correlation map $\vy$. 
    }
    \label{fig:teaser_mfm}
\end{figure}

Our correlational image modeling (\methodname) is a simple yet effective self-supervised pre-training approach. As illustrated in Figure~\ref{fig:teaser_mfm}, we formulate a \textbf{\textit{crop-and-correlate}} pretext task that crops a random image region (\exemplar) from an input image (\context) and predicts the correlation map between the \exemplar and \context. Our \methodname consists of four components: cropping strategy, encoder, decoder, and loss function. In the following, we first introduce correlation operations in Section~\ref{subsec:preliminary}. We subsequently detail each component of \methodname in Section~\ref{subsec:cim}. We finally discuss the relation of our \methodname with the unsupervised visual tracking task in Section~\ref{subsec:comparison}.

\subsection{Preliminary: Correlation Operation}
\label{subsec:preliminary}
Given an \exemplar image $\vz\in\mathbb{R}^{h_z \times w_z \times 3}$ along with a typically larger \context image $\vc \in\mathbb{R}^{
h_c \times w_c \times 3}$, a correlation operation between the \exemplar and \context images is defined as follows:
\begin{equation}\label{eq:corr_ops}
f\left(\vz, \vc\right) = f_\theta(\vz) \star f_\theta(\vc) + \vb \mathds{1},
\end{equation}
where $\star$ denotes a correlation operator and $f_\theta$ is a backbone model to extract corresponding representations. Here, $\vb\mathds{1}$ represents a signal that takes value $\vb$ in every location.  Conceptually, it means that a dense similarity of two sets is measured in a 2D fashion. For instance, in standard Siamese-based trackers~\cite{bertinetto2016fully,valmadre2017end,li2018high,wang2019fast}, $\star$ is normally instantiated as a 2D convolution operator, in which the exemplar feature $f_\theta(\vz)$ takes the role of convolutional kernels, sliding over the spatial region of context feature $f_\theta(\vc)$.
For Transformer-based trackers~\cite{chen2021transformer,xie2022correlation,cui2022mixformer,ma2022unified,song2022transformer}, a cross-attention layer combines information from two images to generate a merged representation, which selectively highlights the hotspots in the \context. 

\begin{figure*}[!ht]
    \centering
    \includegraphics[page=1,width=0.9\linewidth]{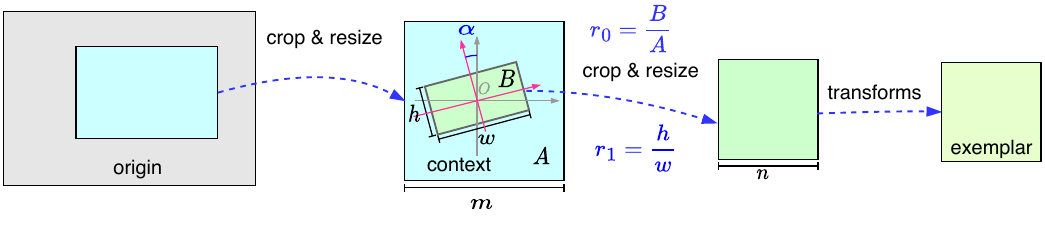}
    \vspace{-10pt}
    \caption{
    The procedure to generate an \textit{exemplar-context} pair for \methodname. We control the scale, shape, and rotation of an \exemplar image by randomly sampling $r_0=\frac{B}{A}$, $r_1=\frac{h}{w}$ and, $\alpha$, where $A$ and $B$ are the areas of cropping region ($h\times w$) and \context image ($m\times m$).
    }
    \label{fig:crop}
\end{figure*}

\subsection{Correlational Image Modeling}
\label{subsec:cim}

\noindent\textbf{Cropping strategy.} To enable effective correlational image modeling for self-supervised visual pre-training, we propose a random cropping strategy to construct \textit{exemplar-context} image pairs. Specifically, as shown in Figure~\ref{fig:crop}, given an original image $\vx\in\mathbb{R}^{ H\times W \times 3}$, we obtain a \context image $\vc\in\mathbb{R}^{ m\times m \times 3}$ with a square shape by first randomly cropping a sub-region followed by a resizing operation. Then, we repeat the \textit{crop-and-resize} process to generate a square \exemplar image $\vz\in\mathbb{R}^{ n\times n \times 3}$ from the \context in consideration of three aspects: scale, shape, and rotation. To control the scale of an \exemplar, we calculate the areas of both the cropping region and \context and compute the scale ratio $r_0$. The shape of the cropping region is determined by the height and width ratio $r_1$. Also, we measure the rotation degree $\alpha$ between the cropping region and \context image. By random sampling the values of $r_0$, $r_1$, and $\alpha$, we can obtain \exemplars with various scales, shapes, and rotations. The corresponding correlation map $\vy \in \{0, 1\}^{m \times m}$ can be derived from the cropping region easily. We further add different transformations to each \exemplar image to increase the data diversity. We will study the effects of different cropping strategies in the experiment section.

\noindent\textbf{Encoder.} The goal of \methodname is to learn useful representations with a backbone model $f_\theta$ in Equation~\ref{eq:corr_ops}, such that $f_\theta$ can be generalized to downstream tasks. Both ViT and CNN architectures can be applied as the encoder for \methodname.
For reliable pre-training, we employ a bootstrap encoder that consists of an online network $\theta$ and a target network $\xi$, which share the same backbone architecture. Given a pair of \exemplar and \context  ($\vz$ and $\vc$), we obtain the corresponding representations:
\begin{equation}\label{eq:encoder}
\vh_z = f_\theta(\vz);
\vh_c = f_\xi(\vc) 
\end{equation}
via the online network $f_\theta$ and target network $f_\xi$, respectively.
The parameters $\xi$ of the target network are updated from the online network $\theta$ with an exponential moving average policy~\cite{lillicrap2015continuous}:
\begin{equation}\label{eq:ema}
\xi = \tau\xi + (1 - \tau)\theta,
\end{equation}
where $\tau \in [0, 1]$ denotes a target decay rate. As a result, the online network $f_\theta$ is responsible for learning the representations to deal with various scales, shapes, and rotations for the \exemplar images. For efficient training, we consider cropping multiple \exemplars for each \context, and all the cropped \exemplars can be grouped together into one forward-backward processing.

\noindent\textbf{Decoder.} To model the correlation between \exemplar and \context images, we design a lightweight cross-attention decoder,   
which is a general form of multi-head attention layer in Transformers~\cite{vaswani2017attention}. 
To be specific, we first project the representations $\vh_z$ and $\vh_c$ to obtain the query, key, and value by linear mappings: $\vq = f_c(\vh_c)+\texttt{PE}; \vk = f_k(\vh_z); \vv = f_v(\vh_z)$. A positional encoding (\texttt{PE}) is added to the query for better decoding. 
The reason why we use the \context as a query rather than the \exemplar is that the output correlation map is of the shape determined by the \context input, not the \exemplar.
Then we can calculate the weighted representation for the \exemplar and \context pair as follows:
\begin{equation}
    \label{eq:ca}
    \va = \texttt{CrossAttention}(\vh_c, \vh_z) = \texttt{Softmax}\biggl(\frac{\vq\trans{\vk}}{\sqrt{d}}\biggl)\vv.
\end{equation}
After that, we compute the correlational representation with \textit{layernorm} and \textit{multilayer perceptron} modules:
\begin{equation}
    \label{eq:de}
    \vh = \vh_c + \va + \texttt{MLP}(\texttt{LN}(\vh_c + \va)).
\end{equation}
Finally, the output correlation map $\widehat{\vy}$ is obtained via a linear predictor and an \textit{upsampling} operation\footnote{https://pytorch.org/docs/stable/generated/torch.nn.Upsample.html}:
\begin{equation}
    \label{eq:pre}
    \widehat{\vy} = \texttt{Upsample}(f_p(\vh)).
\end{equation}

\noindent\textbf{Loss function.} In practice, to optimize the overall \methodname model, we simply minimize a binary cross-entropy loss:
\begin{equation}
    \label{eq:all}
    \mathcal{L}(\widehat{\vy}, \vy) = - \frac{1}{m\times m} \sum^{m\times m}_{i=1} \vy_{i}\log(\widehat{\vy}_i) + (1 - \vy_{i})\log(1-\widehat{\vy}_i),
\end{equation}
between predicted correlation map $\widehat{\vy}$ and ground-truth $\vy$.

\subsection{Relation to Unsupervised Visual Tracking}
\label{subsec:comparison}

Our \methodname is generally related to studies on unsupervised and self-supervised visual tracking~\cite{wang2019unsupervised,wu2021progressive,shen2022unsupervised,lai2020mast,zheng2021learning}. These works explore effective training cues to bypass the need for extensive annotated data for training deep tracking models.
Typically, temporal consistency or correspondence in videos is leveraged as a cue. Several modeling techniques have been proposed, including forward and backward consistency~\cite{wang2019unsupervised}, progressive training~\cite{wu2021progressive}, multi-task learning~\cite{shen2022unsupervised}, and memory augmenting~\cite{lai2020mast}.
Despite the different strategies, all these unsupervised and self-supervised trackers mainly focus on learning task-specific representations for visual tracking from unlabeled videos. Thus, it is infeasible to apply these trackers with temporal modeling on still images, in which such temporal information does not exist.
On the contrary, the goal of our \methodname is to learn generic representations from unlabeled data with the transferable ability to downstream tasks. 
Therefore, we formulate correlational modeling in a more general form and develop it as a useful pretext task for self-supervised visual pre-training. 
\section{Experiments}
\label{sec:exps}
\begin{table}[!h]
\centering
\small
\caption{\textbf{Ablations of cropping strategy} for \methodname with ViT-B/16 on ImageNet-200. 
}
\label{tab:ablation_crop}
\vspace{-8pt}
\begin{subtable}[htp]{0.33\textwidth}
\centering
\small
\caption{\textbf{Crop scale}. 
}
\label{tab:crop_scale}
\resizebox{\textwidth}{!}{
\begin{tabular}{lcc}
\toprule
Scale    & Ratio $r_0$& Top-1 acc (\%) \\ 
\midrule
scratch  & -            & 77.79               \\ 
MoCo v3~\cite{chen2021empirical}  & - & 89.60   \\
MAE~\cite{he2021masked} & -     & 89.03  \\
\midrule
fixed    & $r_0=0.16$         & 87.57   \\
random   & $r_0<0.16$   &     87.25  \\ 
random   & $r_0>0.16$    &     89.39  \\ 
random   & $r_0 \in (0 , 1.0)$   &     \cellcolor{gray!20}\textbf{89.48} \\ 

\bottomrule
\end{tabular}
}
\end{subtable}
\hfill

\begin{subtable}[htp]{0.32\textwidth}
\centering
\small
\caption{\textbf{Crop shape}. 
}
\label{tab:crop_shape}
\resizebox{\textwidth}{!}{
\begin{tabular}{ccc}
\toprule
Shape & Ratio $r_1$ & Top-1 acc (\%) \\ 
\midrule
square    &    1.0   & 89.48               \\
rectangle &$[3/4, 4/3]$  & 89.55 \\
rectangle & $[1/2, 2/1]$  & 89.59 \\
rectangle & $[1/3, 3/1]$  & \cellcolor{gray!20}\textbf{89.70} \\
rectangle & $[1/4, 4/1]$  & 89.66 \\
\bottomrule
\end{tabular}
}
\end{subtable}
\hfill

\begin{subtable}[htp]{0.3\textwidth}
\centering
\small
\caption{\textbf{Rotation}. 
}
\label{tab:rotation}
\resizebox{0.9\textwidth}{!}{
\begin{tabular}{cc}
\toprule
Rotation $\alpha$ & Top-1 acc (\%) \\ 
\midrule
\ang{0}    & 89.70              \\
$[\ang{-45}, \ang{45}]$  & \cellcolor{gray!20}\textbf{89.97} \\
$[\ang{-90}, \ang{90}]$   & 89.91 \\ 
$[\ang{-135}, \ang{135}]$   & 89.19  \\ 
$[\ang{-180}, \ang{180}]$   & 89.07  \\ 
\bottomrule
\end{tabular}
}
\end{subtable}
\vfill

\begin{subtable}[htp]{0.3\textwidth}
\centering
\small
\caption{\textbf{Transformation}. 
}
\label{tab:transformation}
\resizebox{0.90\textwidth}{!}{
\begin{tabular}{ccc}
\toprule
\context & \exemplar & Top-1 acc (\%) \\ \midrule
\xmark     &  \xmark   & 89.97       \\
\cmark     &  \xmark   & 90.01       \\
\xmark     &  \cmark   & \cellcolor{gray!20}\textbf{90.12}     \\ 
\cmark     &  \cmark   & 90.12      \\\bottomrule
\end{tabular}
}
\end{subtable}
\hfill
\end{table}

\subsection{Main Properties}
To understand the unique properties of \methodname, we conduct ablation studies  on ImageNet-200~\cite{van2020scan}, a smaller subset of the ImageNet-1K dataset~\cite{deng2009imagenet}. 
For all ablation experiments, we choose ViT-Base (ViT-B/16) as the default backbone and follow a common setting used in existing works~\cite{bao2021beit,he2021masked}: \textit{300-epoch self-supervised pre-training without labels and 100-epoch supervised end-to-end fine-tuning}, to evaluate the quality of learned representations. 
For a fair comparison, we tailor the resolutions of \textit{context} and \textit{exemplar} as $160\times160$ and $64\times64$, respectively. By default, we crop six \exemplars for each input image (\context), in order to match with the standard $224\times224$ input size.\footnote{For ViT-Base (ViT-B/16) with $16\times16$ patch size, our configuration of one \textit{context} ($160\times160$) with six \textit{exemplars} ($64\times64$) contains 196 image patches in total, which is equivalent to an image with the size of $224\times224$.} More detailed pre-training and fine-tuning recipes are described in the supplementary material. We present our observations as follows: 

\noindent\textbf{Cropping strategy.} We investigate how different cropping strategies will affect our \methodname in self-supervised representation learning. We consider four aspects of cropping \exemplars, \ie, scale, shape, rotation, and transformation:

\textbf{(\romannumeral1) \texttt{Scale}}: As shown in Table~\ref{tab:crop_scale}, we study the scale factor of \textit{exemplars} while keeping the shape and rotation fixed, \ie, square shape and \ang{0} rotation. We first consider cropping with fixed scale ratio $r_0 = \frac{64\times 64}{160\times 160} = 0.16$ and then study three random scale ratio schemes: small scale ($r_0 < 0.16$), large scale ($r_0 > 0.16$), and both small and large scales $r_0 \in (0, 1.0)$. All these entries perform significantly better than the baseline, \ie, training from scratch without pretraining. The random cropping policy that covers both small and large scales $r_0 \in (0, 1.0)$ performs best. This indicates that adding variation on the scale ratio of cropping \exemplar images can help our \methodname to learn better representations.

\textbf{(\romannumeral2) \texttt{Shape}}: In Table~\ref{tab:crop_shape} we further study the crop shape of \exemplars. Given that deep architectures (CNNs and ViTs) are more easily to process rectangle inputs, we extend the square in previous studies to the rectangle with height/width ratio $r_1$. Other non-rectangle shapes (\eg, triangles and circles) are beyond our study. We find that expanding the sampling range of shape ratio $r_1$ as $[1/3, 3/1]$ can boost the performance upon the square entry. However, if a larger range of $[1/4, 4/1]$ is applied, no further performance gain can be obtained. All these experiments suggest that the shape of \exemplars is also a useful factor in our cropping strategy.

\textbf{(\romannumeral3) \texttt{Rotation}}: Table~\ref{tab:rotation} shows the influence of the rotation degree $\alpha$ between
the cropping \exemplars and \context image. We conduct rotation experiments upon the previous best entry in Table~\ref{tab:crop_shape}. 
The optimal sampling range of $\alpha$ is $[\ang{-45}, \ang{45}]$, which means adding a relatively smaller degree of rotation is helpful for our \methodname, while large rotation degree such as $[\ang{-180}, \ang{180}]$ would bring a negative effect. Our consideration of rotation is strategically different from previous predicting image rotations~\cite{gidaris2018unsupervised} in that we treat rotation as a type of augmentation rather than a supervisory signal.
Therefore, a reasonably small degree of rotation can improve our \methodname pre-training.

\textbf{(\romannumeral4) \texttt{Transformation}}: Table~\ref{tab:transformation} studies the influence of data transformations on our \methodname pre-training. We consider random data transformations including horizontal flipping, Gaussian blurring, color jittering, grayscale, and solarization. We can observe that only adding transformations on \exemplars while keeping \context images unaltered works best for our \methodname. This can be explained by our bootstrap encoder design: the online network encode \exemplars while the offline network processes the \context, as a result, \exemplars are more responsible for affecting model training. Note that our transformation for \exemplar-\context pairs is different from existing MV-SSL methods such as MoCo v3~\cite{chen2021empirical} and DINO~\cite{caron2021emerging}, in which two transformed views are conceptually identical, thus can be swapped during training.

\begin{table}[!t]
\centering
\small
\caption{\textbf{Ablations of encoder designs} for \methodname with ViT-B/16 on ImageNet-200. 
}
\label{tab:bootstrap_enc}
\vspace{-8pt}

\centering
\small

\begin{tabular}{ccc}
\toprule
Bootstrap    & Update & Top-1 acc (\%) \\ 
\midrule
scratch  & -            & 77.79               \\ 
\midrule
\xmark  & shared            & 89.91          \\ 
\cmark  & $\xi\rightarrow\theta$  & 89.05 

\\ 
\cmark  & $\theta\rightarrow\xi$ & \cellcolor{gray!20}\textbf{90.12}  \\ 
\bottomrule
\end{tabular}

\end{table}

\begin{figure}[!h]
    \centering
    \includegraphics[page=1,width=1.0\linewidth]{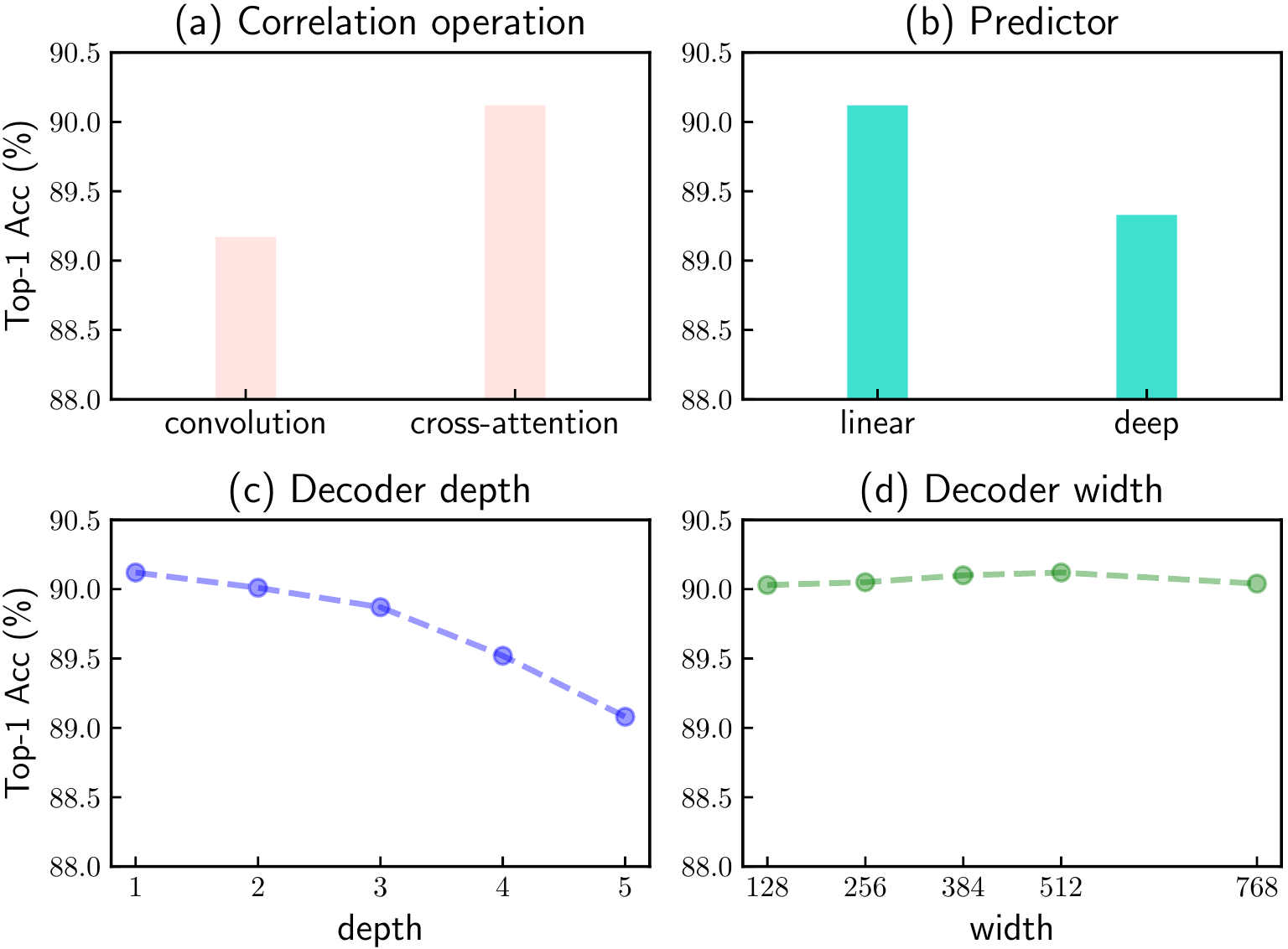}
    \vskip -0.3cm
    \caption{
    \textbf{Ablations of decoder designs} for \methodname with ViT-B/16 on ImageNet-200. 
    }
    \label{fig:dec_design}
\end{figure}

\noindent\textbf{Encoder design.} Our \methodname encoder follows a bootstrap design with \exemplar and \context images encoded by the online network $f_\theta$ and target network $f_\xi$, respectively.
As studied in Table~\ref{tab:bootstrap_enc}, we first notice that learning with a shared encoder can achieve significantly better performance than the training-from-scratch baseline. We further evaluate two bootstrapping designs: 1) $\xi\rightarrow\theta$ training update from \context to \exemplar, and 2) $\theta\rightarrow\xi$ training update from \exemplar to \context. Obviously, we can find that $\theta\rightarrow\xi$ entry performs better than both the shared and $\xi\rightarrow\theta$ entries. This validates the rationality of our bootstrap encoder design, which is also consistent with our finding of transformation in Table~\ref{tab:transformation}: our \methodname benefits more from learning with \exemplars than \context images.

\noindent\textbf{Decoder design.} Our \methodname decoder plays a key role in correlational modeling. We study our decoder designs as follows:

\textbf{(\romannumeral1) \texttt{Correlation operation}}:
Figure~\ref{fig:dec_design} (a) compares two correlation operations commonly used in existing deep tracking models. As we introduced in Section~\ref{subsec:preliminary}, when applying convolution as the correlation operation, \exemplars features are served as the kernels and convolve with \context features, in which local correlations are computed in each kernel window. Differently, as formulated in Equation~\ref{eq:ca}, a cross-attention layer models global correlation between \exemplar and \context images. The cross-attention entry can yield up to 1\% improvement over the convolution entry. This suggests that modeling global correlation is better for \methodname.

\textbf{(\romannumeral2) \texttt{Predictor}}: In Figure~\ref{fig:dec_design} (b), we study the network design of the final predictor. A simple linear layer followed by an \texttt{Unsample} operation works well for our \methodname. While a deep predictor with three deconvolution layers cannot 
bring further gain but degrade \methodname training. This suggests a lightweight predictor may force \methodname to better representations in the encoder, while a heavy predictor is more specialized for predicting accurate correlation maps in the decoder but less relevant for representation learning.

\textbf{(\romannumeral3) \texttt{Depth}}: Figure~\ref{fig:dec_design} (c) varies the decoder depth (number of cross-attention layers). Interestingly, a \textit{single} cross-attention layer works best for our \methodname training. Adding more layers brings no training gain for correlation modeling, similar to our observation in previous predictor designs.  

\textbf{(\romannumeral4) \texttt{Width}}: Figure~\ref{fig:dec_design} (d) studies the decoder width (number of heads in each cross-attention layer). We set 512-d by default, which performs well for our \methodname. Increasing or decreasing the layer width does not cause significant accuracy improvement or degradation. The decoder depth is less influential
for improving representation learning for our \methodname.

Overall, our \methodname decoder is lightweight. It has only one cross-attention layer with a width of 512-d and a linear layer for final prediction.  As such, our \methodname is efficient in model pre-training.

\noindent\textbf{Loss function.} We study the influence of different loss functions for our \methodname optimization in Table~\ref{tab:loss_fuc}. Given that our \methodname predicts binary correlation maps, we compare typical loss functions for dense predictions, including cross-entropy (CE), balanced cross-entropy (BCE)~\cite{xie2015holistically}, mean squared error (MSE), and Focal loss~\cite{lin2017focal}. A standard cross-entropy performs best for our correlation modeling. This property is dramatically different from deep visual tracking models~\cite{bertinetto2016fully,valmadre2017end,wang2021multiple} and related unsupervised trackers 
~\cite{wang2019unsupervised,wu2021progressive,shen2022unsupervised,lai2020mast,zheng2021learning}, which clearly benefit from proper dense objectives. This can be explained by the difference in task goals between \methodname and visual tracking: our \methodname focuses on learning transferable representations by correlation modeling, whereas deep visual trackers demand task-specific representations in favor of better dense predictions. 

\begin{table}[!t]
\centering
\small
\caption{\textbf{Ablations of loss fuctions} for \methodname with ViT-B/16 on ImageNet-200. 
}
\label{tab:loss_fuc}
\addtolength{\tabcolsep}{-2pt}
\vspace{-8pt}
\begin{tabular}{l|cccccc|c}
\toprule
Loss Function & CE &  BCE &  MSE & Focal  \\ \midrule

Top-1 acc (\%) & \cellcolor{gray!20}\textbf{90.12}  & 89.28 & 87.68      & 77.35    \\
\bottomrule
\end{tabular}
\end{table}

\begin{table}[!t]
\centering
\small
\caption{\textbf{Comparisons with visual tracking works} with ViT-B/16 on ImageNet-200. 
}
\label{tab:vis_tracking}
\addtolength{\tabcolsep}{-2pt}
\vspace{-8pt}
\begin{tabular}{l|ccccccc|c}
\toprule
Method  & Scratch &  SiamFC &  SiamRPN & TransTrack & \methodname  \\ \midrule

Top-1 (\%) &77.79  & 89.09  & 89.02  & 89.54 & \cellcolor{gray!20}\textbf{90.12} \\

\bottomrule
\end{tabular}
\end{table}

\begin{table*}[!t]
\centering
\caption{\textbf{ImageNet-1K top-1 fine-tuning accuracy} of self-supervised models using ViT-S/16 and ViT-B/16 as the encoder. 
All entries are on an image size of $224\times224$. 
We use the actual processed images/views to measure the effective pre-training epochs~\cite{zhou2022image}. 
Scratch indicates the supervised baseline in~\cite{touvron2021training}. 
$^\dag$ denotes results are reproduced using the official code.}
\label{tab:vit_results}
\addtolength{\tabcolsep}{-2pt}
\vspace{-8pt}
\begin{tabular}{lccccccc} 
\toprule
Method & Pre-train Data  & Pretext Task  & Tokenizer  & Epochs & ViT-S & ViT-B \\ \midrule
Scratch~\cite{touvron2021training} & - & -               & -            & -                  & 79.9      & 81.8      \\ \midrule
MP3~\cite{zhai2022position}    & IN-1K &  Jigsaw              & -                      & 100       & -      & 81.9      \\ \midrule
MoCo v3~\cite{chen2021empirical} & IN-1K & MV-SSL               & -                       & 1200       & 81.4      & 83.2      \\
DINO~\cite{caron2021emerging}    & IN-1K & MV-SSL              & -                      & 1600       & 81.5      & 82.8      \\ \midrule
BEiT~\cite{bao2021beit}    & IN-1K+DALL-E  & MIM             & dVAE                      & 300       & 81.3      & 82.9      \\
SimMIM~\cite{xie2021simmim}$^\dag$  & IN-1K   & MIM            & -                     & 300       & 80.9      & 82.9     \\
MAE~\cite{he2021masked}$^\dag$     & IN-1K    & MIM           & -                     & 300       & 80.6      & 82.9     \\ 
\midrule 
\methodname     & IN-1K        & CIM       & -                       & 300       & 81.6      &  83.1     \\ \bottomrule
\end{tabular}
\end{table*}

\subsection{Comparisons with Visual Tracking Models}
Our \methodname is inspired by the correlation modeling in supervised visual tracking models. Our proposed cropping strategy can generate useful \exemplar-\context pairs that are also suitable for training supervised visual tracking models. We train three representative trackers using ViT-B/16 as the backbone: SiamFC~\cite{bertinetto2016fully}, SiamRPN~\cite{li2018high}, and TransTrack~\cite{chen2021transformer}, with generated \exemplar-\context pairs on ImageNet-200. Following the same pre-training and fine-tuning setting in previous ablation studies, we evaluate the quality of learned representations, as summarized in Table~\ref{tab:vis_tracking}. We can observe that: (1) Owing to the \exemplar-\context pairs generated by our cropping strategy, all three trackers can learn good representations that perform better than the scratch baseline. (2) Based on SiamFC, SiamRPN introduces an additional detection head for bounding box prediction, which brings no performance gain. (3) TransTrack works better than both SiamFC and SiamRPN. This is due in large part to the beneficial global correlation modeling provided by cross-attention layers, in comparison with local correlations computed by convolution operations in SiamFC and SiamRPN. (4) Our \methodname clearly surpasses these visual tracking works, showing the advantages of our encoder and decoder designs for effective correlational modeling.

\subsection{Comparisons with Previous SSL Methods}
\label{exps:sota}

Our \methodname is a general framework that can learn meaningful representations for both ViT and CNN architectures, unlike state-of-the-art methods such as MAE~\cite{he2021masked}. 

\noindent\textbf{ViT.} In Table~\ref{tab:vit_results} we first compare the fine-tuning results of ViT-S/16 and ViT-B/16 models with self-supervised pre-training on ImageNet-1k. Following previous works~\cite{bao2021beit,he2021masked}, we fine-tune ViT-S/16 for 200 epochs, and ViT-B/16 for 100 epochs.
More detailed pre-training
and fine-tuning configurations are described in the supplementary material. 
Compared with previous MV-SSL works~\cite{chen2021empirical,caron2021emerging}, such as MoCo v3~\cite{chen2021empirical}, our \methodname can achieve highly comparable performances (83.1 vs. 83.2), while enjoying significantly fewer epochs of pre-training (300 vs. 1200). 
Compared with previous MIM works~\cite{bao2021beit,he2021masked,xie2021simmim}, using the same 300 epochs of pre-training, our \methodname can achieve better performances with both ViT-S/16 and ViT-B/16 models.

\begin{table}[ht]
\centering
\caption{\textbf{ImageNet-1K top-1 fine-tuning accuracy} of self-supervised models using ResNet-50 as the encoder. $^\dag$ denotes results are reproduced using the official code.
}
\label{tab:resnet50}
\addtolength{\tabcolsep}{-2pt}
\begin{subtable}[t]{0.45\textwidth}
\centering
\vspace{-8pt}
\label{tab:resnet50_2}
\resizebox{\textwidth}{!}{
\begin{tabular}{lccc}
\toprule
Method & Pretext Task   & Epochs & Top-1 acc (\%) \\ \midrule
\multicolumn{4}{l}{\emph{Fine-tuning for 100 epochs}}            \\
RSB A3~\cite{wightman2021resnet}  & -  & -       & 78.1           \\
SimMIM~\cite{xie2021simmim}$^\dag$   & MIM    & 300       & 77.7           \\
\methodname    & CIM   & 300       & 78.6           \\ \midrule
\multicolumn{4}{l}{\emph{Fine-tuning for 300 epochs}}     \\
RSB A2~\cite{wightman2021resnet}  &-  & -       & 79.8           \\ 
SimSiam~\cite{chen2021exploring}   & MV-SSL     & 400       & 79.1           \\
MoCo v2~\cite{chen2020improved}   & MV-SSL & 400       & 79.6           \\
SimCLR~\cite{chen2020simple}  & MV-SSL & 800       & 79.9           \\
BYOL~\cite{grill2020bootstrap}  & MV-SSL & 400       & 80.0           \\
SwAV~\cite{caron2020unsupervised}  & MV-SSL & 600       & 80.1           \\ 
SimMIM~\cite{xie2021simmim}$^\dag$ & MIM  & 300       & 79.5           \\
\methodname     & CIM  & 300       & 80.1           \\ \bottomrule
\end{tabular}
}
\end{subtable}
\end{table}

\noindent\textbf{ResNet-50.} We further demonstrate that our \methodname can effectively pre-train the classic ResNet architecture. During pre-training, we simply apply the same ViT pre-training configurations for ResNet-50. To evaluate the pre-trained representations, we generally follow the state-of-the-art vanilla ResNet “training-from-scratch” recipe in RSB~\cite{wightman2021resnet}. We present detailed fine-tuning settings in the supplementary material. The evaluation results compared to the state-of-the-art methods are summarized in Table~\ref{tab:resnet50}. Due to the architectural difference between ViT and CNN models, we observe performance degeneration of some MIM and MV-SSL pre-training methods, such as SimMIM~\cite{xie2021simmim}, MoCo v2~\cite{chen2020improved}, and SimSiam~\cite{chen2021exploring}. Compared with the best MV-SSL method, SwAV~\cite{caron2020unsupervised}, our \methodname is faster (300 vs. 600). 

Overall, our \methodname is a simple yet effective approach that can perform on par or better than existing MV-SSL and MIM methods with both ViT and ResNet models.

\begin{figure*}[!h]
    \centering
    \includegraphics[page=1,width=1.0\linewidth]{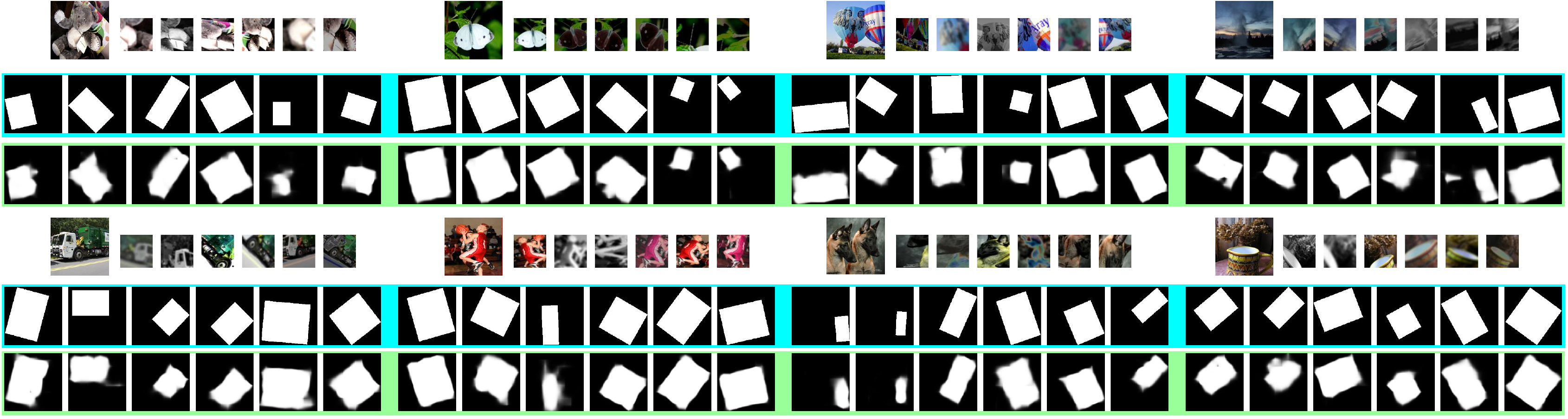}
    \vskip -0.3cm
    \caption{
    \textbf{Visualization} of \exemplar-\context images in company with both \textcolor{aqua}{ground-truth} and \textcolor{guppiegreen}{predicted correlation} maps for \methodname.
    }
    \label{fig:cim_vis}
\end{figure*}

\subsection{Transfer Learning on Semantic Segmentation}
To evaluate the transferability of the pre-trained representations by our \methodname, we further conduct end-to-end fine-tuning on the ADE20K~\cite{zhou2019semantic} semantic segmentation benchmark.
Following the same setup in BEiT~\cite{bao2021beit}, we fine-tune the pre-trained ViT-B/16 model as the backbone in UperNet~\cite{xiao2018unified} for 160K iterations, with an input resolution of $512\times512$. As summarized in Table~\ref{tab:seg}, our \methodname can achieve highly competitive performance compared with other representative self-supervised learners. This demonstrates the effectiveness of our proposed \textit{crop-and-correlate} pretext task in learning transferable representations.

\begin{table}[!t]
\centering
\caption{\textbf{ADE20K semantic segmentation} of ViT-B/16 models. 
}
\label{tab:seg}
\addtolength{\tabcolsep}{-3pt}
\vspace{-8pt}
\resizebox{0.45\textwidth}{!}{
\begin{tabular}{lccc}
\toprule
Method     & Pre-train Data & Pretext Task & mIoU (\%) \\ \midrule
Supervised~\cite{touvron2021training} & IN-1K w/ labels & -& 45.3 \\ \midrule
MoCo v3~\cite{chen2021empirical}    & IN-1K & MV-SSL & 47.2 \\
DINO~\cite{caron2021emerging}       & IN-1K & MV-SSL & 46.8 \\ \midrule
BEiT~\cite{bao2021beit}       & IN-1K+DALL-E & MIM & 47.7 \\
MAE~\cite{he2021masked}        & IN-1K & MIM & 48.1 \\ \midrule
\methodname        & IN-1K & CIM & 48.1 \\ \bottomrule
\end{tabular}
}
\end{table}

\subsection{Robustness Evaluation}

We further evaluate the robustness of our models on six benchmarks that cover the challenges of adversarial attacks,
common corruption, and out-of-distribution. For adversarial attack, we evaluate the adversarial examples on ImageNet-A~\cite{hendrycks2021natural}, along with generated examples by FGSM~\cite{goodfellow2014explaining} and PGD~\cite{madry2017towards} attackers on ImageNet-1K validation set. For data corruption, we test corrupted images on ImageNet-C~\cite{hendrycks2021many}. In terms of out-of-distribution input, we consider images with 
distribution shifts from ImageNet-R~\cite{hendrycks2021many} and ImageNet-Sketch~\cite{wang2019learning}.  Specifically, we directly evaluate the models fine-tuned on original ImageNet-1K (ViT-B/16 in Table~\ref{tab:vit_results} and ResNet-50 in Table~\ref{tab:resnet50}) without further fine-tuning on each robustness validation set. The results are summarized in Table~\ref{tab:robustness}. For both ViT and ResNet architectures, our \methodname consistently outperforms the state-of-the-art self-supervised learners for model robustness.

\begin{table}[h]
\centering
\small
\caption{\textbf{Robustness evaluation on six robustness benchmarks.} We report top-1 accuracy except for IN-C which uses the mean corruption error (mCE). The original ImageNet top-1 fine-tuning results are also appended for reference. The best results are in \textbf{bold}, and the second best results are \underline{underlined}.}
\label{tab:robustness}
\addtolength{\tabcolsep}{-2pt}
\vspace{-8pt}
\resizebox{.48\textwidth}{!}{%
\begin{tabular}{lccccccc}
\toprule
\multirow{2}{*}{Method} & \multicolumn{6}{c}{Robustness Benchmarks} & \multirow{2}{*}{Orig.} \\ \cmidrule(lr){2-7}
                        & FGSM  & PGD  & IN-C ($\downarrow$) & IN-A & IN-R & IN-SK &                        \\ \midrule
\multicolumn{8}{l}{\emph{ViT-B/16 model results}} \\  
Scratch~\cite{wightman2021resnet}                 & 46.3      & 21.2     & \textbf{48.5}     & 28.1     & 44.7     & 32.0      & 81.8                       \\ 
MAE~\cite{he2021masked}                     & 38.9      & 11.2     & 52.3     & 31.5    & 48.3     & 33.8      & \underline{82.9}                       \\

\methodname                     & \textbf{47.4}      & \textbf{22.7}     & \underline{49.3}     & \textbf{30.3}     & \textbf{48.6}     & \textbf{35.3}      & \textbf{83.1}                       \\ \midrule 
\multicolumn{8}{l}{\emph{ResNet-50 model results}} \\  
Scratch~\cite{wightman2021resnet}                 & \textbf{20.2}      & \textbf{3.4}     & 77.0     & 6.6     & 36.0     & 25.0      & \underline{78.1}                       \\ 

SimMIM~\cite{xie2021simmim}                 & 16.8      & 2.1     & 77.0     & 5.7     & 34.9     & 24.2      & 77.7                      \\
\methodname                     & \underline{19.4}      & \underline{2.5}     & \textbf{73.5}     & \textbf{8.5}     & \textbf{37.4}     & \textbf{27.2}      & \textbf{78.6}                       \\ \bottomrule
\end{tabular}%
}

\end{table}

\subsection{Visualization}
In Figure~\ref{fig:cim_vis}, we visualize the \exemplar-\context images generated by our proposed cropping strategy on the ImageNet-1K validation set. The predicted correlation maps are obtained via the ViT-B/16 model pre-trained on the ImageNet-1K train set using our \methodname in Section~\ref{exps:sota}. No further pre-training or fine-tuning is conducted.
We can observe that these predicted correlation maps match closely with the corresponding ground-truth correlations, under various scales, shapes, rotations, and transformations.
The results demonstrate the effectiveness of self-supervised correlation modeling in our \methodname for unseen data.  More visualized examples are provided in the supplementary material.
\section{Conclusion}
\label{sec:conclude}
In this work, we present \methodname, a novel pretext task for self-supervised visual pre-training. Unlike existing MV-SSL and MIM approaches, \methodname considers correlation modeling in visual tracking as a useful pre-training paradigm. We build a generic self-supervised correlational modeling framework by proposing three unique designs, including a cropping strategy, bootstrap encoder, and correlation decoder. Extensive experiments on transfer learning and robustness evaluation with visual recognition tasks show that our \methodname can efficiently and effectively learn meaningful representations from unlabeled images.

\noindent\textbf{Acknowledgement.}
This study is supported under the RIE2020 Industry Alignment Fund – Industry Collaboration Projects (IAF-ICP) Funding Initiative, as well as cash and in-kind contribution from the industry partner(s). It is also supported by Singapore MOE AcRF Tier 2 (MOE-T2EP20120-0001).

{\small
\bibliographystyle{ieee_fullname}
\bibliography{egbib}
}

\clearpage
\appendix
\section{Appendix}
In the supplementary material, we provide the detailed pre-training and fine-tuning recipes in Section~\ref{sec:supp_impl}. Section~\ref{sec:supp_visual} provides more qualitative visualization for \exemplar-\context images and predicted correlation maps.

\subsection{Implementation Details}
\label{sec:supp_impl}

\noindent\textbf{Pre-training.}
Table~\ref{tab:supp_pretrain_setting} summarizes the pre-training settings for vanilla ViT and ResNet-50 models. All experiments are conducted on 8 A100 GPUs for both ViT and ResNet-50 models. 
Our \methodname is \emph{general} across architectures that the configurations are \emph{shared} by different architectures, without specialized tuning. 

\noindent\textbf{Fine-tuning.}
Table~\ref{tab:supp_finetune_vit_setting} and Table~\ref{tab:supp_finetune_r50_setting} summarize the fine-tuning settings for vanilla ViT and ResNet-50 models, respectively. The configurations for ViT are \emph{shared} across models. The configurations for ResNet-50 basically follow~\cite{wightman2021resnet}, using the AdamW optimizer following~\cite{fang2022corrupted}.

\noindent\textbf{Semantic segmentation on ADE20K.}
Following the configurations in BEiT~\cite{bao2021beit}, we fine-tune UperNet~\cite{xiao2018unified} using AdamW as the optimizer for 160K iterations with a batch size of 16. The input resolution is $512\times512$, and we use single-scale inference.
Following the common practice of BERT~\cite{devlin2019bert} fine-tuning in NLP~\cite{pruksachatkun2020intermediate}, we initialize all segmentation models using model weights after supervised fine-tuning on ImageNet-1K as suggested in BEiT~\cite{bao2021beit}.

\begin{table*}[h]
\centering
\small
\caption{\textbf{Pre-training settings for vanilla ViT-S/16, ViT-B/16 and ResNet-50 models on ImageNet-200 and ImageNet-1K.} Note that we adopt the \emph{same} pre-training configurations across different architectures without further parameter tuning.}
\label{tab:supp_pretrain_setting}
\begin{tabular}{ll}
\toprule
Configuration \hspace{20pt}                  & Value \\ \midrule
Optimizer \hspace{20pt}                      & AdamW~\cite{loshchilov2017decoupled}      \\
Pre-training epochs \hspace{20pt}            & 300      \\
Peak learning rate \hspace{20pt}             & 2.4e-3      \\
Batch size \hspace{20pt}                     & 4096      \\
Weight decay \hspace{20pt}                   & 0.05      \\
Optimizer momentum \hspace{20pt}             & $\beta_1,\beta_2=0.9,0.95$~\cite{chen2020generative}      \\
Learning rate schedule \hspace{20pt}         & Cosine decay      \\
Warmup epochs \hspace{20pt}                  & 40      \\
Gradient clipping \hspace{20pt}              & 1.0      \\
Dropout~\cite{srivastava2014dropout} \hspace{20pt}                        & \ding{55}      \\
Stochastic depth~\cite{huang2016deep} \hspace{20pt}               & \ding{55}      \\
LayerScale~\cite{touvron2021going} \hspace{20pt}                     & \ding{55}      \\
Data augmentation \hspace{20pt}              & RandomResizedCrop      \\
Pos. emb. in Transformer layers \hspace{20pt} & 1-D absolute pos. emb.~\cite{dosovitskiy2020image}    \\
Patch size \hspace{20pt}                     & 16      \\
Pre-training resolution of \context  image       & 160      \\ 
Pre-training resolution of \exemplar  image        & 64      \\
Number of \exemplars \hspace{20pt}       & 6      \\ 
\bottomrule
\end{tabular}
\end{table*}

\begin{table*}[h]
\centering
\small
\caption{\textbf{Fine-tuning settings for vanilla ViT-S/16 and ViT-B/16 on ImageNet-200 and ImageNet-1K.} We fine-tune ViT-S/16 for 200 epochs, and ViT-B/16 for 100 epochs. All other hyper-parameters are the same.}
\label{tab:supp_finetune_vit_setting}
\begin{tabular}{ll}
\toprule
Configuration \hspace{20pt}                 & Value \\ \midrule
Optimizer \hspace{20pt}                     & AdamW~\cite{loshchilov2017decoupled}      \\
Fine-tuning epochs \hspace{20pt}            & 200 (S), 100 (B)      \\
Peak learning rate \hspace{20pt}            & 9.6e-3      \\
Layer-wise learning rate decay~\cite{bao2021beit} \hspace{20pt} & 0.8~\cite{clark2020electra}      \\
Batch size \hspace{20pt}                    & 2048      \\
Weight decay \hspace{20pt}                  & 0.05      \\
Optimizer momentum \hspace{20pt}            & $\beta_1,\beta_2=0.9,0.999$      \\
Learning rate schedule \hspace{20pt}        & Cosine decay      \\
Warmup epochs \hspace{20pt}                & 5      \\
Loss function \hspace{20pt}                 & Cross-entropy loss \\
Gradient clipping \hspace{20pt}             & \ding{55}      \\
Dropout~\cite{srivastava2014dropout} \hspace{20pt}                        & \ding{55}      \\
Stochastic depth~\cite{huang2016deep} \hspace{20pt}              & 0.1      \\
Mixup~\cite{zhang2017mixup} \hspace{20pt}                         & 0.8      \\
Cutmix~\cite{yun2019cutmix} \hspace{20pt}                        & 1.0      \\
Label smoothing~\cite{szegedy2016rethinking} \hspace{20pt}               & 0.1      \\
Random augmentation~\cite{cubuk2020randaugment} \hspace{20pt}           & 9 / 0.5      \\
Patch size \hspace{20pt}                    & 16      \\
Fine-tuning resolution \hspace{20pt}        & 224      \\
Test resolution \hspace{20pt}               & 224      \\ \bottomrule
\end{tabular}
\end{table*}

\begin{table*}[t]
\centering
\small
\caption{\textbf{Fine-tuning settings for vanilla ResNet-50 on ImageNet-1K.} The hyper-parameters generally follow~\cite{wightman2021resnet}, except that we adopt the AdamW optimizer following~\cite{fang2022corrupted}.}
\label{tab:supp_finetune_r50_setting}
\begin{tabular}{lcc}
\toprule
Configuration \hspace{20pt}                        & 100 epoch FT \hspace{20pt}          & 300 epoch FT         \\ \midrule
Optimizer \hspace{20pt}                     & \multicolumn{2}{c}{AdamW~\cite{loshchilov2017decoupled}}                     \\
Peak learning rate \hspace{20pt}            & \multicolumn{2}{c}{12e-3}                     \\
Layer-wise learning rate decay~\cite{bao2021beit} \hspace{20pt} & \multicolumn{2}{c}{\ding{55}}                          \\
Batch size \hspace{20pt}                    & \multicolumn{2}{c}{2048}                      \\
Weight decay \hspace{20pt}                  & \multicolumn{2}{c}{0.02}                      \\
Learning rate schedule \hspace{20pt}        & \multicolumn{2}{c}{Cosine decay}              \\
Warmup epochs \hspace{20pt}                 & \multicolumn{2}{c}{5}                         \\
Loss function \hspace{20pt}                 & \multicolumn{2}{c}{Binary cross-entropy loss} \\
Gradient clipping \hspace{20pt}             & \multicolumn{2}{c}{\ding{55}}      \\
Dropout~\cite{srivastava2014dropout} \hspace{20pt}                       & \multicolumn{2}{c}{\ding{55}}                          \\
Stochastic depth~\cite{huang2016deep} \hspace{20pt}              & \multicolumn{2}{c}{\ding{55}}                          \\
Mixup~\cite{zhang2017mixup} \hspace{20pt}                         & \multicolumn{2}{c}{0.1}                       \\
Cutmix~\cite{yun2019cutmix} \hspace{20pt}                        & \multicolumn{2}{c}{1.0}                       \\
Label smoothing~\cite{szegedy2016rethinking} \hspace{20pt}               & 0.1 \hspace{20pt}                   & \ding{55}                     \\
Repeated augmentation~\cite{berman2019multigrain,hoffer2019augment} \hspace{20pt}         & \ding{55} \hspace{20pt}                      & \ding{51}                     \\
Random augmentation~\cite{cubuk2020randaugment} \hspace{20pt}           & 6 / 0.5 \hspace{20pt}               & 7 / 0.5              \\
Fine-tuning resolution \hspace{20pt}        & 160 \hspace{20pt}                   & 224                  \\
Test resolution \hspace{20pt}               & \multicolumn{2}{c}{224}                       \\
Test crop ratio \hspace{20pt}              & \multicolumn{2}{c}{0.95}                      \\ \bottomrule
\end{tabular}
\end{table*}

\begin{figure*}[!h]
    \centering
    \includegraphics[page=1,width=1.0\linewidth]{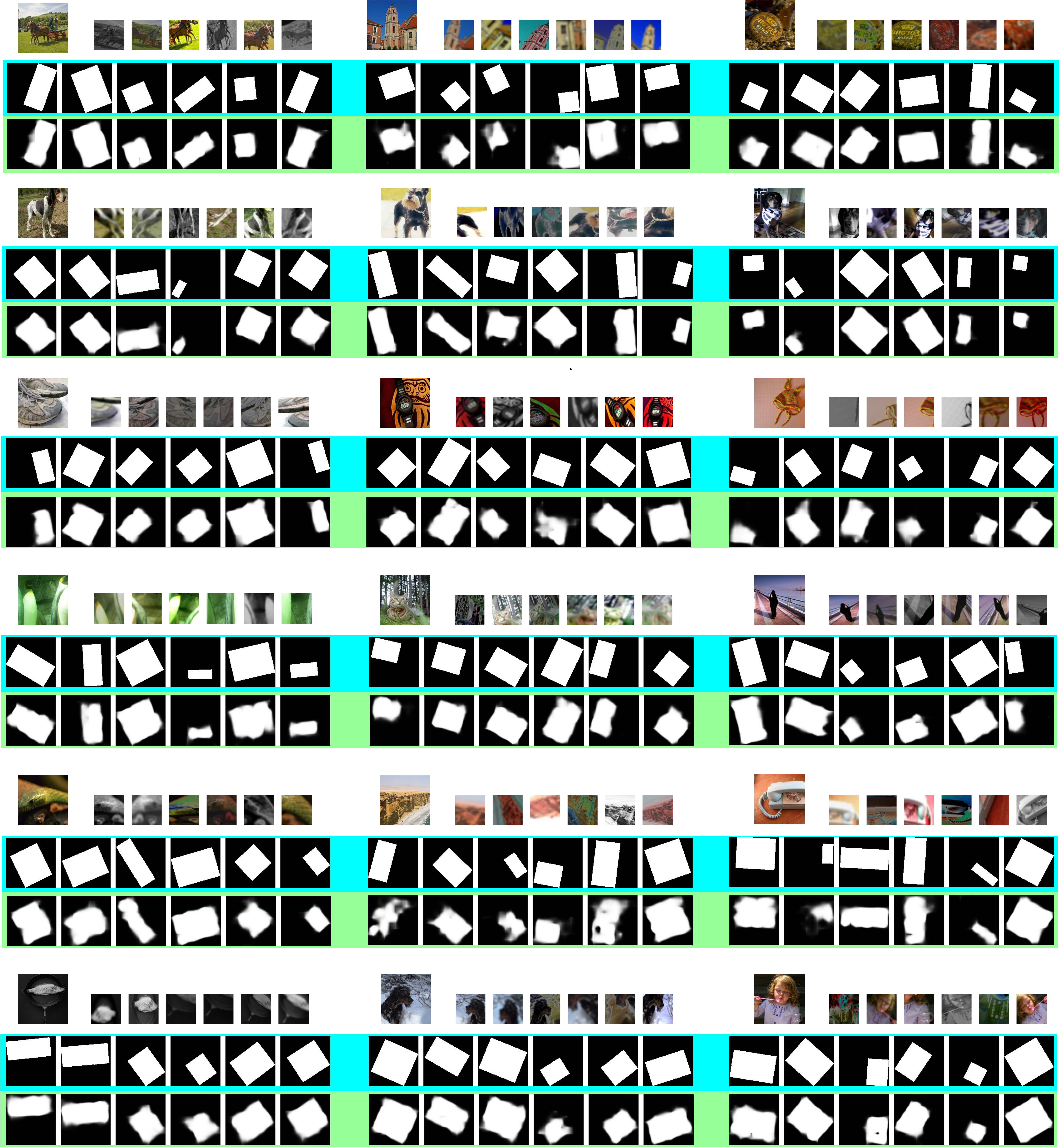}
    \caption{
    \textbf{Visualization} of \exemplar-\context images in company with both \textcolor{aqua}{ground-truth} and \textcolor{guppiegreen}{predicted correlation} maps for \methodname.
    }
    \label{fig:cim_vis_more}
\end{figure*}

\subsection{More Visualization}
\label{sec:supp_visual}

We provide more qualitative visualization of \exemplar-\context images together with both ground-truth and predicted correlation maps for \methodname in Figure~\ref{fig:cim_vis_more}, using unseen ImageNet-1K \emph{validation} images.

\end{document}